# Unraveling the Versatility and Impact of Multi-Objective Optimization: Algorithms, Applications, and Trends for Solving Complex Real-World Problems

Noor A. Rashed
*Department of Computer Science*
*University of Technology*
*- Iraq*
cs.20.62@grud.uotechnology.edu.iq

Yossra H. Ali
*Department of Computer Science*
*University of Technology*
*- Iraq*
yossra.h.ali@uotechnology.edu.iq

Tarik A. Rashid
*Department of Computer Sciences & Engineering*
*University of Kurdistan*
*Hewler, Iraq*
tarik.ahmed@ukh.edu.krd

A.Salih
*Department of Artificial Intelligence Technology*
*Northumbria University*
*London, UK*
dr.salisonn@outlook.com

**Abstract**

*Multi-Objective Optimization (MOO) techniques have become increasingly popular in recent years due to their potential for solving real-world problems in various fields, such as logistics, finance, environmental management, and engineering. These techniques offer comprehensive solutions that traditional single-objective approaches fail to provide. Due to the many innovative algorithms, it has been challenging for researchers to choose the optimal algorithms for solving their problems. This paper examines recently developed MOO-based algorithms. MOO is introduced along with Pareto optimality and trade-off analysis. In real-world case studies, MOO algorithms address complicated decision-making challenges. This paper examines algorithmic methods, applications, trends, and issues in multi-objective optimization research. This exhaustive review explains MOO algorithms, their methods, and their applications to real-world problems. This paper aims to contribute further advancements in MOO research. No singular strategy is superior; instead, selecting a particular method depends on the natural optimization problem, the computational resources available, and the specific objectives of the optimization tasks.*

## 1- Introduction

Decision-makers are faced with the daunting task of trying to achieve several objectives at once everyone is competing for attention and resources. From optimizing the costs for production and environmental impacts to balancing stakeholders' needs and wants, the demand for appropriate optimization techniques has never been higher. Fortunately, the arsenal of optimization algorithms has expanded, offering diverse approaches to address these nuanced challenges. Several real-world search and optimization issues are organically formulated as nonlinear programming problems with multiple objectives. Due to a lack of suitable solution strategies, these problems were artificially converted into single-objective problems and solved [1]. Such problems produce a set of trade-off optimal solutions, known as the Pareto-optimal solutions, rather than a single solution. It becomes pertinent to look for as many Pareto-optimal solutions as possible if it is possible to demand more than one. Algorithms that seek to solve cases where there are several times varying objectives that are usually conflicting are referred to as Multi-Objective Optimization (MOO) or Multi-Objective Evolutionary Algorithms (MOEAs) [2]. In general, the problem of this kind of optimality concerns only one criterion function that should be maximized or minimized. However, in practical situations, numerous goals must be considered simultaneously [3]. The idea of MOO is thus to find the 'optimum' which will best meet all the mutually exclusive objectives. Pareto optimal solutions offer the best opportunities with the least undesirable trade-offs that favor the higher number of benefits over the number of costs. The Pareto set, or Pareto front, is a set of solutions to define the best result according to the Pareto principle [4]. This review provides an effective overview of MOO algorithms and how they function in practice, methodology, and applications to various real-life problems. Discussions on the Pareto principle and the importance of trade-off type research for decision-making are made. These methodologies, including Non-dominated Sorting Genetic Algorithm II (NSGA-II) and Multi-Objective Fitness-Dependent Optimizer (MOFDO) algorithm, enable decision-makers to understand the complexities of trade-offs between multiple goals. However, due to their computational requirements and potential to become massive, they still require significant attention in practice [3], [5]. Among all those approaches, Evolutionary Algorithms (EA) have become prominent as they are capable of identifying the particular trade-off space and provide a range of Pareto-optimal solutions, besides the decomposition-based methods, and how they can be used to solve different kinds of problems will be discussed. On the other hand, algorithms are more suitable for solving complex problem landscapes, such as Multi-Objective Simulating Annealing (MOSA) with Multi-Objective Invasive Weed Optimization Algorithm (MOIWOA) but suffer from certain problems that can be seen in real-world scenarios: their reliance on parameter tuning and resource-intensive computations [6]. Meanwhile, Indicator-based methods, as represented by the Island Multi-Indicator Algorithm (IMIA) — provide consistently high performance when it comes to solving complex and Pareto fronts and optimizing solutions with complex geometries. However, they are very sensitive to indicators selection and specific characteristics of problems, thus requiring a careful approach towards their application [7]. Case studies from the real world show how well MOO systems work when making hard decisions. There are many ways to use it, such as in supply chain management, portfolio optimization, treatment planning, resource sharing, and other areas [8], [9]. Also, every method has its strengths and weaknesses, and the choice of method depends on many factors, such as the nature of the optimization problem, the computational resources available, and the specific objectives of the optimization tasks. MOO can be applied to a range of different sorts of problems, and this reveals how it will be able to assist people who are in a position where they have to make decisions. Then, it details the methods and uses of MOO and the current research trends and problems that are coming up in the field. The most important aspects related to the context of MOO are discussed, including the issues of how to handle the uncertainty, about objectives and restrictions of decision-makers, how to solve the high-dimensional problems, and how to avoid the "curse of dimensionality."

This paper intends to contribute positively to knowledge in the field of MOO by presenting principles and methodologies used, examples of applications and the challenges faced in the field. It means that by engaging with the specifics of MOO algorithms and how they can be used in practice we will be able to better understand the nuances of the corresponding processes. This paper endeavors to fill the gap between theory and practice, proposing guidelines that can inform decision-making processes across various domains.

The outline of this paper is as follows: Section 2 introduces the MOO methods. Section 3 presents the applications of MOO algorithms. Section 4 provides a discussion and analysis of the work. Finally, the conclusions are given in Section 5.

## 2- Multi-Objective Optimization

An unconstrained continuous Multi-Objective Problem (MOP) can be formally described without compromising generality as follows [6]:

$$minimize F(Q) = (f_1(Q_1), f_2(Q_2), \ldots, f_n(Q_n))^\top \qquad (1)$$

$$subject\ to\ Q \in \Omega$$

where the variable vector is denoted as $Q = (Q_1, Q_2, \ldots, Q_n)^\top$, $\Omega$ represents the variable space, $n$ is represented by the variable number, the constructed of $n$ real-valued objective functions denoted as $F: \Omega \to R^n$, and the objective space is represented by $R^n$. The attainable objective can be defined as the set $\{F(Q) \mid Q \in \Omega\}$.

### Definition 1: Domination

$p = (p_1, \ldots, p_n)^\top, v = (v_1, \ldots, v_n)^\top \in R^n$ Are pair of vectors, $p$ dominates $v$ if $p_j \le v_j$ for every $j \in \{1,2,3,\ldots,n\}$. At least one exists $i \in \{1,2,3,\ldots,n\}$ satisfying $p_i < v_i$, which can be represented as $p > v$.

### Definition 2: Pareto optimal solution

A point $K^* \in \Omega$ is named as a Pareto optimal to (1) only if pointless $K \in \Omega$ satisfying $F(K)$ dominates $F(K^*)$.

### Definition 3: Pareto optimal Set (*PS*)

The *PS* can be described as $PS = \{K^* \in \Omega \mid \neg \exists K \in \Omega: F(K) > F(K^*)\}$; it represents all Pareto optimal solutions.

### Definition 4: Pareto Front (*PF*)

Corresponding with the meaning of PS, the Front of Pareto (PF) is characterized as follows: P$F$= $\{F(K) \mid K \in PS\}$, displaying all objective space Pareto optimum solutions.

MOEAs have three MOP goals: (1) good convergence, (2) good variety, and (3) good coverage, which means it can cover the complete PF.

When attempting to identify MOO, utilizing a collection of solutions from several methods is common. Figure 1 displays the structure of the methods used to balance competing goals.

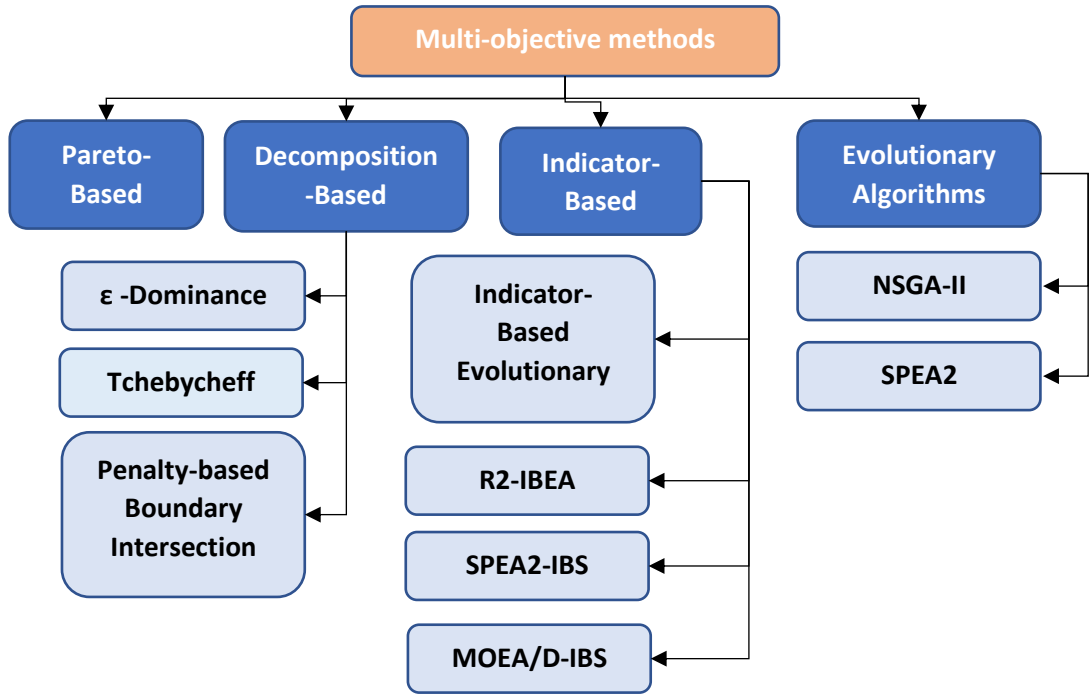

Figure 1: The multi-objective methods

## 2.1 Pareto-based Methods

Pareto-based methods, which are also called Pareto optimization or MOO, are ways to solve problems where different goals are essential. The Italian economist Vilfredo Pareto developed the idea of Pareto efficiency, which is how it got its name [7]. Pareto-based techniques seek Pareto-optimal alternatives. No alternative answer can advance one goal without damaging another [8]. This is to give people who have to make decisions choices trade-offs between different goals, as shown in Figure 2.

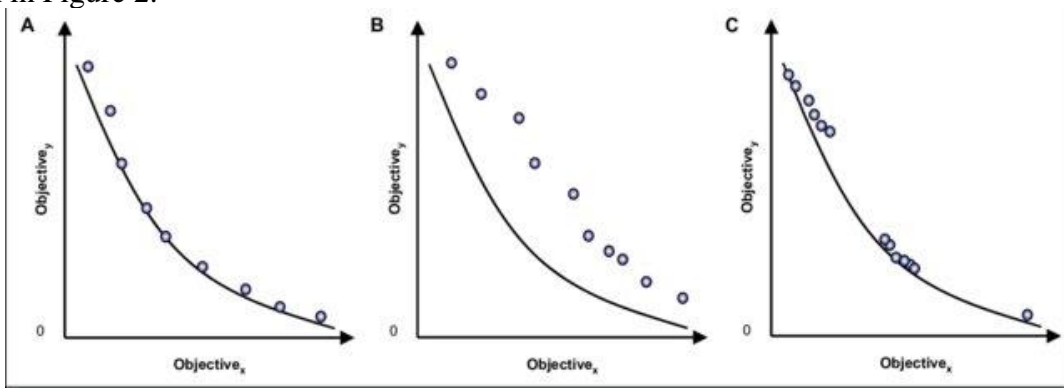

Figure 2: The sets of solutions generated by a multi-objective Pareto technique [7].

The Pareto dominance idea is a famous tool in Pareto-based techniques. It ranks solutions based on how well they meet several criteria. Dominant solutions are at least as good as other solutions and better in at least one way. These methods can find Pareto-optimal answers by comparing and contrasting them repeatedly. Pareto-based solutions could be useful in engineering [10], finance [11], logistics [12], and many other fields where choices must be made while trying to balance competing goals.

## 2.2 Decomposition-based Methods

The challenges of MOO can be tackled with the help of Decomposition-Based Methods (DBMs), a family of optimization techniques. These techniques break down the original MOP into simpler optimization issues with a single target [13]. By combining the results of these smaller issues, an estimate of the Pareto front can be made. DBMs come in a few versions, but all work the same way. The weighted sum method is a popular approach whereby the MOP is converted into a single-objective problem by allocating different weights to the objectives in each sub-problem [14]. The shape of the final Pareto front is modified by weights that, in turn, define an optimum point that favors the target achievements [15], [16], [17]. The explicit MOP can be broken down into individual subproblems, each linked to a single individual and optimized using data from its neighboring subproblems. Decomposition-based multi-objective optimization relies on the principle that optimal solutions to neighboring subproblems should be near one other and that any knowledge gained from one subproblem should be valuable for optimizing another subproblem [18].

### 2.2.1 The ε -Dominance Method

It is a type of dominant relation that allows for some degree of choice amongst different solutions, and is frequently employed when many solutions share similar objective values and must be differentiated by a fixed threshold [19]. In ε-Dominance, a way out it is said that solution A "dominates" solution B if it is at least as good as solution B in all goals and better than solution B by a certain amount in at least one objective. The parameter ε is a small positive number that shows how much someone can handle or wants something [20]. ε-Dominance can be used to divide a set of solutions into various Pareto fronts. Each front is made up of solutions that are not dominated by any other solution in the set. In MOO, this allows answers to be ranked in a more complex and flexible way [21]. This strategy chooses one objective function and constrains the others to make the problem a single-objective problem. The different $\varepsilon$ results represent optimal Pareto answers. The prescribed format of this approach is given below.

$$\min F(X) = \{f_i(y), \dots, f_n(y)\} \quad \text{subject to}: g(y) <=> b \text{ and } x \geq 0 \quad (2)$$

$$\min F(X) = \{f_i(y), \dots, f_n(y)\} \quad \text{subject to}: g(y) <=> b, f_j \leq \varepsilon_j,$$

$$j \neq i, j = 1, \dots, n, \text{and } y \geq 0 \quad (3)$$

If the objective function is max, the constraint is $f_j(y) \geq \varepsilon_j$. Selecting the $\varepsilon$ is important since this option affects answers so much. So, the selected $\varepsilon$ must be in the range of $f_j^{\min} \leq \varepsilon_j \leq f_j^{\max}$ for each objective function [18].

### 2.2.2 The Tchebycheff Approach

The Tchebycheff approach (also known as the Tchebycheff method or the Tchebycheff scalarization) is widely applied in MOO. Considering the worst-case results across all goals, this DBM seeks to identify the optimal compromise option [22]. The Tchebycheff decomposition breaks a MOP into scalar optimization subproblems as follows:

$$\min_{x \in \Omega} g^{tch}(F(y) \mid w, u^*) = \max_{1 \leq i \leq m} \{w_i(f_i(y) - u_i^*)\} \quad (4)$$

where $\sum_{i=1}^{m} w_i = 1$ and $w_i \geq 0$ is the vector of weights for a subproblem, and $u^* = (u_1^*, \ldots, u_m^*)$ with $u_i^* < \min\{f_i(y) \mid y \in \Omega\}$ is a perfect objective vector. The traditional Tchebycheff decomposition does not explore subproblem objective functions' geometric properties. This is the first study to look into the geometric feature of Tchebycheff decomposition subproblem objective functions.

**Theory 2.1**: Let $u^* = (u_1^*, \ldots, u_m^*)$ be a perfect objective vector, and $w = (w_1, \ldots, w_m)$ be assigned as a positive weight vector.

If the objective vector is given as: $F(y) = (f_1(y), \ldots, f_m(y))$ is in the line

$$L_1: w_1(f_1(y) - u_1^*) = \cdots = w_m(f_m(y) - u_m^*) \quad (5)$$

Then

$$g^{tch}(F(y) \mid w, z^*) = \frac{w^T(F(y) - u^*)}{m} \quad (6)$$

The traditional Tchebycheff decomposition does not explore the geometric characteristics of objective functions in subproblems. This study is the first to investigate the geometric feature of objective functions in the Tchebycheff decomposition subproblem. With the Tchebycheff approach, a weight vector or reference point is introduced to the MOO problem, turning it into a single-objective problem. The reference point is a user-defined in the goal space that symbolizes the desired solution, and the weight vector denotes the relevance or priority assigned to each objective [23]. The highest weighted departure from the reference point among all objectives is then used in the Tchebycheff scalarization formula to determine a scalar value for each candidate solution [21]. Thus, Tchebycheff scalarization seeks the least favorable outcome concerning the objective problem function. The purpose is to eliminate the maximum deviations from the optimization target. Next, the Tchebycheff optimization locality is reached when values of scalarization tend to be minimal to solve the problem. By applying this procedure, leaders can first rank their favorite targets and then strike a compromise that suits them best [24]. The method is less rigid and simplified enough to be adaptable for different cases. Maintenance is predominantly about MOO. This is a multi-functional device that can be used for weighing and a lot more things. Pareto optimization offers a different view where the ranking of alternatives is considered, and an array of Pareto-optimal solutions can then be developed.

### 2.2.3 The Penalty-based Boundary Intersection Approach

In multi-objective optimization, the search is generally directed towards the Pareto the front using the Penalty-primarily based Boundary Intersection (PBI) approach. To reap a well-rounded investigation of the goal area, it combines the concept of boundary intersection with penalty functions [25].

The PBI method frames optimization as a multi-objective minimization hassle [26]. This approach results in a penalty function that draws solutions too close to the Pareto front, which searches for alternatives more consistent with the Pareto front [27].

The objective features and penalty terms are combined using a secularizing characteristic in the PBI approach. The optimization system is directed toward the threshold of the Pareto the front using the secularizing characteristic. By increasing penalties for solutions that stray from the boundary intersection websites, it finds an appropriate balance between absolutely exploring the objective space and convergent shifting closer to the Pareto front [28]. The PBI method can be written as follows:

$$g^{PBI}(x \mid \lambda, \theta) = d_1 + \theta d_2$$
$$d_1 = \frac{\|(f(x) - z^*)^T \lambda\|}{\|\lambda\|}$$
$$d_2 = \left\| f(x) - \left(d_1 \frac{\lambda}{\|\lambda\|} + z^*\right) \right\| \quad (7)$$

where $\theta(\theta \geq 0)$ denotes the penalty value. $d_1$ denotes the distance between the ideal point $z^*$ and the projection of $x$ on the search direction line, $d_2$ denotes the perpendicular distance to the search direction line. Both the distances should be as small as possible [26].

The PBI approach encourages algorithms to explore the Pareto front better by including penalty capabilities and boundary intersection notions. It encourages answers to converge closer to the actual Pareto front while preserving solution variety [29].

The PBI approach has proven to be useful in numerous multi-objective optimization situations, yielding top-quality Pareto-front answers [30].

For complex issues concerning numerous targets and selection variables, decomposition-primarily based processes have been significantly used in MOO. By breaking down the problem and successfully navigating the alternate-off area, they provide an opportunity for traditional EAs.

## 2.3 Indicator-based Methods

Several researchers have turned to indicator-based multi-objective optimization strategies to find Pareto's most appropriate solutions to multi-goal optimization problems. By including signs that quantify the exceptional diversity of solutions, those methods try to strike a stability between convergence, which means identifying answers close to the Pareto front, and variety, which means retaining a numerous organization of answers [31]. Some well-known MOO approaches are listed below, all of which rely on indicators.

### 2.3.1 Indicator-Based Evolutionary Algorithm

Indicator-Based Evolutionary Algorithm (IBEA) is an algorithm that employs indications to direct the search. It uses a metric, like hyperactive volume or spacing, to assess the range and depth of the possible answers. Using the solution's impact on the indicator's total value, IBEA calculates a fitness score for it. Using this fitness assignment method, IBEA promotes both Pareto front exploration and solution diversity [32].

### 2.3.2 R2-Indicator-Based Evolutionary Algorithm

R2-IBEA is an enhancement of IBEA that adds a 2D indicator. The R2 value quantifies the coverage of the goal area via the answers. R2-IBEA can discover components of the objective space underserved by way of gift answers by using the R2 indicator. It limits the quest for unique websites to grow the number of solutions. This procedure is executed most effectively offline [33].

### 2.3.3 Strength Pareto Evolutionary Algorithm Version Two: Indicator-Based Selection

The benefits of Strength Pareto Evolutionary Algorithm Version Two (SPEA2) and Indicator-Based Selection (IBS) have been combined in SPEA2-IBS. The dominance relationships between potential solutions are measured by a strength number, while an indicator is used to help make the final selection. Since indicator values measure quality and diversity, higher indicator values are more likely to be chosen. This combined approach boosts convergence and solution variety in SPEA2-IBS [34].

### 2.3.4 Indicator-Based Selection for Multi-Objective Evolutionary Algorithm-based Decomposition

A new version called IBS for MOEA-based Decomposition (MOEA/D-IBS), IBS is added to the MOEA/D algorithm. A MOEA/D-IBS indicator, like hyper-volume, is used to measure the quality and variety of answers for each sub-problem [35]. It promotes the search to zero in on areas that better cover the Pareto front, achieved by selecting and reproducing solutions with higher indicator values [36].

These MOO approaches rely on performance indicators to direct the search process and give a framework for doing so. These techniques use indicators to efficiently probe the Pareto front and keep a range of options available to indicate the compromises necessary to achieve competing goals [37]. The indication and method will be chosen based on the type of problem and the user's preferences [38].

### 2.4 Evolutionary Algorithms

In a single simulation session, the population method of EAs quickly finds many Pareto-optimal solutions. Because of this, Evolutionary Multi-objective Optimization (EMO) study and use have become very popular in the last ten years [2]. Figure 3 shows the flowchart of the EAs, which can deal with goals that are at odds with each other; they are often used to solve MOPs. These algorithms are based on the way evolution works in nature. They use techniques like selection, reproduction, and mutation to find the best answers in a space with multiple goals [39].

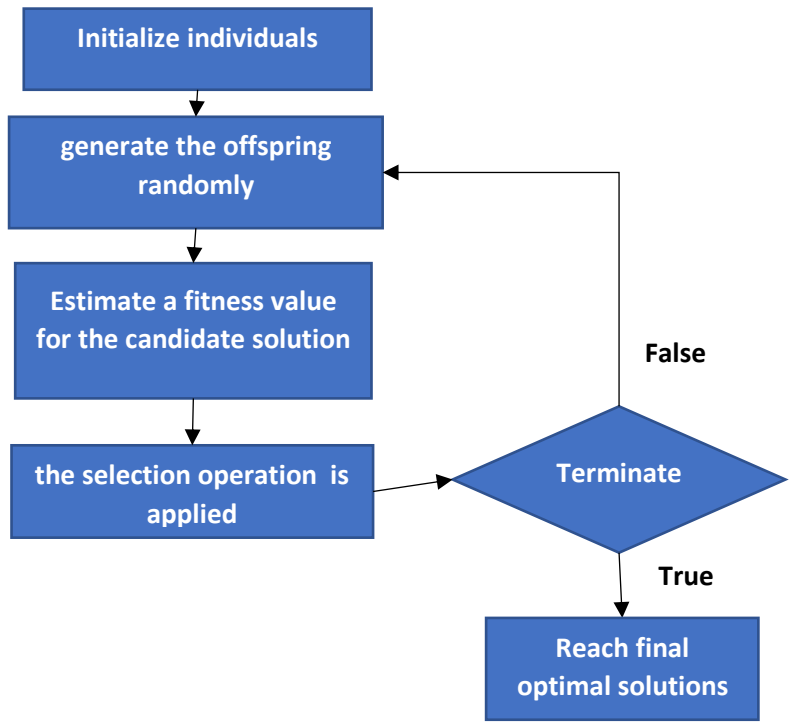

Figure 3: An example flowchart of the EAs [40].

### 2.4.1 Non-dominated Sorting Genetic Algorithm II

NSGA-II is a famous EA used to optimize more than one thing. It saves a list of all the possible answers and uses non-dominated sorting and crowding distance to help the search find several Pareto-optimal solutions [41]. It uses a fast-sorting method called "non-dominated sorting" to put solutions in order based on their dominance relationships. This lets the algorithm keep the variety of solutions. NSGA-II has been widely used and demonstrated effective performance for many MOO tasks [42].

Customers can create offspring (N), and then the worst N people within the discern and offspring population are eliminated by using rank and crowding distance. The smallest rank will live on to the following era, and the important rank (this is, if all individuals with this rank live on, the populace size might be at the least N but less than N if all individuals no longer enter the next era) will simply have its people with the largest crowding distances live to tell the tale to make sure the subsequent population length (O) [43].

### 2.4.2 Strength Pareto Evolutionary Algorithm 2

The well-known algorithm SPEA2 also uses a population-based method but uses a different strategy to keep various solutions [44]. It uses a concept called "strength" to measure the quality of solutions and combines this with a density estimation method to choose the best solutions for reproduction. SPEA2 has done a good job of solving MOPs with many goals [45]. NSGA-II and SPEA2 are not the only evolutionary methods for MOO, but there are many more. MOEA/D is one of these [46], IBEA, and many others. Each algorithm has its strengths and flaws, and how well it works depends on how the optimization problem is set up [7].

## 3- Applications using Multi-Objective Algorithm

Many problems in the real world have many different, sometimes conflicting goals and a large search space. Conflicting goals lead to Pareto-optimal compromises. No trade-off is better without preference knowledge. However, accurate approaches cannot solve huge and complex search spaces. Thus, effective optimization solutions must address both issues [2], [47].

In this section, some works that use multi-objective algorithms in their applications will be shown:

1. Marina Khoroshiltseva *et al.* [48] designed and built a new energy-efficient fixed daylight around the outside window of a Madrid apartment building. MOO algorithm combined Harmony search and Pareto-based approaches to optimize the shading scheme for a social housing complex in Madrid. The resulting methodology provided here focuses on energy efficiency. The best way to design effective shading devices is when the performance of the proposed solution depends on several conflicting objectives.
2. Chao Ni *et al.* [49] built a Multi-Objective Feature Selection (MOFS) with ways to enhance it. One optimization intention is to select as few tendencies as possible related to fee analysis. Another objective is to utilize Pareto-based multi-objective optimization Algorithms (PMAs) to enhance the overall efficiency of software defect prediction SDP models. The analysis examined the impact of different PMAs on MOFES and determined that NSGA-II performed well on all datasets.
3. Erfan Babaee Tirkolaee *et al.* [50] provide reliable and enduring solutions that enhance the efficiency of supply chain operations, create transportation routes, reduce greenhouse Gas (GHG) emissions, and improve customer satisfaction. This study presented a variant of Bi-Objective Mixed-Integer Linear Programming (BOMILP) with two objectives that aim to minimize the overall cost while utilizing the MOSA and NSGA-II algorithms to identify Pareto solutions that maximize distributed reliability. This revolutionary solution to docking and sustainable supply chain management should be seen. The BOMILP version aims to minimize pricing frequency and maximize supply chain reliability by considering transportation route planning, GHG emissions, and consumer satisfaction. Statistical testing is used to assess the effectiveness of the proposed strategies in addressing real-world problems. They have found that response methods can also produce superior solutions, and NSGA-II is the most effective tool.
4. Jesus Guillermo Falc *et al.* [51] used IMIA to overcome the limitation of IB-MOEAs-based multi-objective optimization problems with irregular Pareto-front geometry. It connects the MOEAs. IMIA outperforms the original IB-MOEA and some current MOEAs by their detection biases. The invariance of the Pareto-front makes IMIA ideal for MOPs with complex Pareto-front geometries.
5. Jie Hou *et al.* [42] presented a study that employed the NSGA-II to enhance the crossover and mutation probabilities through reinforcement learning. This led to improved joint growth and runtime results for the manipulator, surpassing other conventional optimization methods. The total runtime was reduced by 19.26%, enhancing work efficiency.
6. Kuihua Huang *et al.* [52] proposed a competition and cooperation approach based on a robust Pareto-evolution strategy to reduce the makespan and Total Energy Consumption (TEC) in distributed heterogeneous permutation flow shop scheduling problems. The heuristic information from each solution divides the population into statistical tests. Speed increases the convergence of lost solutions and winners. Problem-based initialization, local search, and energy conservation strategies also reduce makespan and total energy consumption.
7. Mingjing Wang *et al.* [53] suggest an approach based on MOEA, named MOEA/D-HHL, that combines deconstruction with Harris Hawks Learning (HHL) for medical device learning. The MOEA/D-HHL algorithm was effectively used to analyze clinical data on lupus nephritis and pulmonary hypertension. The algorithm achieved a normalized mutual information score of 0.9652 and an adjusted rand index score of 0.9749 for lupus nephritis. For pulmonary hypertension, the normalized mutual information score was 0.9686, and the adjusted rand index score was 0.9742. The analytical results demonstrate that the MOEA/D-HHL algorithm surpasses other techniques in treating Pulmonary hypertension of lupus nephritis. The statistical analysis indicates that all indicators have predictive capabilities. The proposed HHL approach in MOEA/D is a more resilient alternative to the existing system developed for the medical device class.
8. Saúl Zapotecas-Martínez *et al.* [54] developed a box restriction of the multi-objective continuous optimization problem and used EA based on the Lebesgue measure. The constant property of the continuous multi-objective optimization problems is exploited privately by the proposed method of continuous problems with approximate Pareto sets, and this property is effective when solving. The results demonstrate the high competitiveness of the presented approach, which in many cases provided the evolutionary criteria-based algorithms used in their investigation of multi-objective success problems.
9. Amir M. Fathollahi-Fard *et al.* [55] developed a context-dependent and resilient optimization method to handle uncertainty in logistical and service aspects of home healthcare and to plan resources for home healthcare effectively. Their advanced optimization model for sustainable and efficient home healthcare logistics and services incorporates a heuristic technique based on the Lagrangian relaxation notion. Three heuristic algorithms are employed to tackle the problem. Heuristic algorithms symmetrically route chemists and patients. The epsilon constraint approach and Lagrangian relaxation theory yield very efficient Pareto-based solutions rapidly. A comprehensive analysis demonstrates that both the multi-objective optimization version and the proposed heuristic approach are environmentally friendly and feasible. An analysis is conducted to examine certain sensitivities to provide managerial insights for developing sustainable and robust home healthcare services.
10. Seyed Jalaleddin Mousavirad, *et al.* [56] proposed a multi-objective approach to improve image quality and reduce image file size by using a quantization table in JPEG image compression, which can help in mobile device energy efficiency. This study also offers secular methods based on Pareto-principles. The authors incorporate the proposed method under metaheuristic algorithms, such as Enhanced Multi-Objective Genetic Algorithm (EnMOGA), Enhanced Multi-Objective Practice Swarm (EnMOPS), and Enhanced Multi-objective Non-dominated Sorting Genetic Algorithm-II (EnNSGA-II). The results show that the EnMOGA and EnMOPS prioritized objective functions 6 and 7, in 13 secularization processes, respectively. EnNSGAII outperformed EnNSGAIII in 10 excess measurements among 13 Pareto-based algorithms. The Wilcoxon signed-rank test is used to statistically evaluate the performance of the algorithm and perform sensitivity analysis.

11. Erfan Babaee Tirkolaee *et al.* [57] suggested a multi-objective mixed-integer linear programming model for supply chain decisions involving multiple periods, echelons, and products. The authors simultaneously mitigate human peril, pollution, and expense. Multi-objective Gray Wolf Optimizer (MOGWO) and NSGA-II discover Pareto optimal solutions by solving the model. The MOGWO algorithm generates a COVID-19 mask closed-loop supply chain network more effectively. By 25% in Pareto solution dispersion and 2% in solution quality, MOGWO surpasses NSGA-II.

12. Erfan Babaee Tirkolaee *et al*. [58] suggested a sustainable garbage collection model. This work introduced a mixed-integer linear programming model to decrease costs, environmental emissions, and operational variability while optimizing public satisfaction. The authors developed a hybrid multi-objective optimization technique to address the challenge properly. The objectives aim at minimizing the set costs, approximate the impact on environment, influence the concern level, and guarantee that variations in workload will not be an issue. To solve the problem with the highest level of effectiveness, the authors introduced the concept of a mixed multi-objective optimization algorithm identified as MOSA with MOIWOA.

13. Jaza M. Abdullah *et al.* [59] proposed a Multi-Objective Fitness-Dependent Optimizer (MOFDO). The authors evaluate how appropriate the algorithm is for benchmark issues that are artificial and for real-life situations encountered in engineering. The impacts show that the effects verify that MOFDO is better than the alternative optimization approaches in terms of convergence and answer diverse. Moreover, MOFDO presents a vast array of feasible options that choice-makers can select from, hence acting as a tool for addressing rational ingenuity design issues.

14. Chnoor M. Rahman et al. [60] designed an overall performance-based learner conduct gadget to solve engineering optimization problems with multiple goals by introducing a new multi-objective optimization set of rules called the Multi-Objective Based learner Performance (MOBP) primarily based behavior algorithm. As for the evolution, the algorithm was used with five real tasks in engineering, and the case was different in the comparison with three other multi-goal ones. The result shows that the advised set of rules outperformed the others in both accuracy and variety.

**4- Discussions and Analysis**

Complex issues with conflicting aims require multi-objective algorithms. Instead of focusing on one goal, these algorithms create solutions that optimize numerous criteria concurrently. This section examines the effects of applying multi-objective algorithms to engineering, finance, and data analysis. Figure 4 shows the rank of researchers used in this paper.

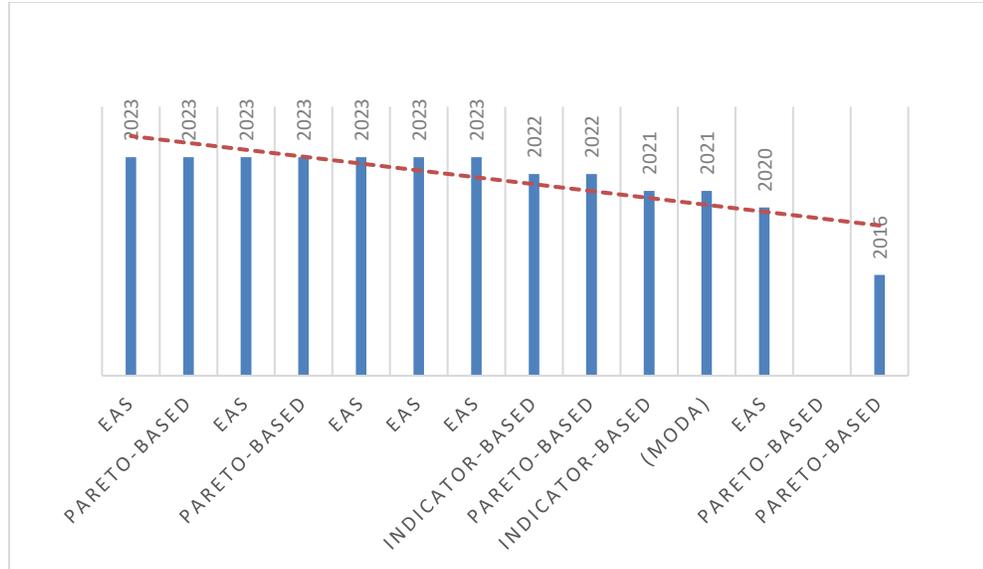

Figure 4: The rank of researchers

Table 1 discusses the pros and cons of these algorithms and how they can assist decision-makers in balancing competing priorities.

**Table 1: A Comparative Analysis of Related Works**

| Ref. | Application | Method | Results | Pros | Cons |
|---|---|---|---|---|---|
| [4] | Solve a variety of engineering optimization problems | (NSGA-II), (MOWCA) and (MODA). | Surpassed the others in accuracy and diversity. | It offers a range of algorithms suitable for different engineering optimization problems. | It requires extensive parameter tuning and adaptation for specific optimization problems to get optimal performance. |
| [48] | Shading design | Pareto-based | It is useful for energy-efficient shading devices with many conflicting goals | It allows for the consideration of multiple conflicting objectives, leading to comprehensive solutions. | It requires significant computational resources due to the complexity of Pareto-based optimization. |
| [49] | Software defect prediction | Pareto-based | NSGA-II outperforms other PMAs on MOFES. | It is Effective in identifying trade-offs between conflicting objectives in software defect prediction. | It requires scalability and convergence for large-scale software systems. |
| [50] | The dependable pollution routing | EAs | Minimizes the total cost and maximizes | It addresses real-world logistics | It requires extensive |
| | problem involves selecting cross-docks using multi-objective optimization. | | supply reliability while considering transportation route design | challenges with consideration for multiple objectives. | parameter tuning and problem-specific customization. |
| **[51]** | Overcome the limitations of IB-MOEAs Address the multi-objective optimization problems that involve continuous variables and are subject to box constraints. | Indicator-Based | Because of its Pareto-front shape invariance, IMIA can also optimize MOPs with intricate Pareto-front geometries that outperform state-of-the-art indicator | It Offers robust performance for optimization problems with complex Pareto fronts. It Provides accurate assessments of solution quality in multi-objective optimization. | It is sensitive to the choice of indicators and problem characteristics. It May face challenges with computational efficiency for high-dimensional problems. |
| **[52]** | Resolve the dispensed heterogeneous permutation drift and keep the scheduling hassle | EAs | Split the population, accelerate loser convergence and winner discovery, and reduce makespan and TEC (Total Elapsed Completion Time). | It provides efficient scheduling solutions for heterogeneous permutation drift keep scheduling problems. | It struggles with scalability for large-scale scheduling instances. |
| **[53]** | Medical machine learning | EAs | All indicators are predictive and more stable for a developing medical machine learning system. | It delivers stable and predictive solutions for medical machine-learning applications. | It requires extensive computational resources for training complex machines. |
| **[54]** | Joints and manipulator runtime | EAs | Has decreased the manipulator running time by 19.26%. | It Effectively optimizes the runtime performance of robotic manipulators. | It requires fine-tuning of algorithm parameters for specific robotic systems. |
| **[55]** | Home Healthcare Logistics: COVID-19 Pandemic Response | Pareto-based | Multi-objective optimization models are efficient and practical and analyze a range of sensitivities to supply managerial insights for sustainable and convenient home healthcare services. | It addresses complex logistics challenges during the COVID-19 pandemic with consideration for multiple objectives. | The challenges in real-world implementation are due to the dynamic nature of the pandemic response. |
| **[56]** | Improve image quality and minimize image file size | Pareto-based | JPEG image compression increases image quality based on the quantization table and reduces image file size | It Enhances image quality while reducing file size, suitable for various image processing applications. | It leads to the loss of fine details in highly compressed images. |
| **[57]** | COVID-19 Sustainable Mask Closed-Loop Supply Chain Network | EAs | Reduce costs, pollution, and human danger concurrently | It Addresses sustainability challenges in the supply chain network during the COVID-19 pandemic. | Requires coordination among various stakeholders and regulatory compliance. |
| **[58]** | Permanent garbage collection arc routing Trouble | Pareto-based | Minimize expenses, environmental emissions, and workload variance while maximizing citizenship satisfaction | It Offers comprehensive solutions for optimizing garbage collection operations considering multiple objectives. | It is challenging due to the complexity of urban logistics and municipal regulations. |
| **[59]** | Conflicting multi-objective optimization | EAs | MOFDO optimizes best. MOFDO offers various viable engineering design solutions to decision-makers. | Fast convergence Provide good diversity | It requires extensive parameter tuning and adaptation for specific optimization problems to get optimal performance |

This section emphasizes that many real-world situations feature a complex search space and many competing objectives. Effective optimization algorithms that can handle both challenges are needed in such a situation [37]. One such method that helps manage challenging decision-making scenarios is the use of multi-objective algorithms.

Segment maintained by bringing up posted works that use multi-objective algorithms in their designs. For instance, Marina Khoroshiltseva *et al.* [48] presented a multifaceted optimization methodology for shading design to address the challenge of creating novel energy-efficient fixed daylight systems that enclose the external windows of a residential building in Madrid. The algorithm combined Harmony search with Pareto-based optimization to identify the most significant trade-offs.
Another case in point is the work by S. S. Rathore *et al.* [49], who optimized the design of a solar-powered water pumping system using a multi-objective optimization method. The program discovered optimal compromise options that improved the system's efficiency while reducing costs.

This section aims to show how valuable multi-objective algorithms can be applied to real-world issues with numerous competing goals and intricate search spaces. These algorithms can aid decision-makers in establishing the optimal balance between competing objectives.

## 5- Conclusion

This paper offers a comprehensive review to explore multi-objective algorithms and their various uses. It explores optimization methods for difficult issues with conflicting goals. The results show the importance of these algorithms in numerous fields and their potential to change decision-making.

This review paper starts by explaining multi-objective optimization's essential notions and how these algorithms handle conflicting goals and discover the algorithms like NSGA-II, SPEA2, and MOEA/D strengths, which efficiently explore Pareto fronts and capture varied solutions.

The applications section demonstrated the actual implementation of these algorithms in several fields. Multi-objective algorithms excel in finding optimal solutions for several objectives in engineering, finance, healthcare, and environmental management. In portfolio optimization, medical treatment planning, sustainable resource allocation, and their ability to find Pareto-optimal solutions usher in a new era of informed decision-making. The studied literature emphasizes the need to continuously enhance and adapt multi-objective algorithms to different problem landscapes. Combined with new technologies like machine learning and artificial intelligence, these algorithms open up new optimization possibilities.

This paper gives the readers a bird's-eye view of MOO algorithms, methodologies, and applications. It helps scholars and practitioners use MOO to make smart decisions in complex real-world situations by combining knowledge from diverse domains. A fascinating world of computational brilliance and practical significance in applications of multi-objective algorithms has been explored.